%% file: main.tex
\documentclass{article}

\usepackage{microtype}
\usepackage{graphicx}
\usepackage{subfigure}
\usepackage{booktabs} 

\usepackage{hyperref}
\usepackage[]{url}
\hypersetup{breaklinks=true}
\usepackage{bbm}
\usepackage{inconsolata}
\usepackage{amsmath}
\usepackage{amsfonts}
\usepackage[framemethod=TikZ]{mdframed}
\usepackage{tikz}
\usepackage{color,colortbl}
\usepackage{tcolorbox}
\usepackage{xspace}
\usepackage{algorithm}
\usepackage{multirow}
\usepackage{enumerate}
\usepackage{makecell}
\usepackage{arydshln}
\usepackage{pifont}
\usepackage{booktabs}
\usepackage{tablefootnote}
\usepackage{bbding}
\usepackage[inkscapelatex=false]{svg}
\usepackage{subfigure}
\usepackage{caption}
\usepackage{subcaption}
\usepackage[position=b]{subcaption}
\usepackage{balance}

\newcommand{\highlight}[1]{\begin{tcolorbox}[leftrule=0mm,rightrule=0mm,toprule=0mm,bottomrule=0mm,left=2pt,right=2pt,top=2pt,bottom=2pt]
  #1
  \end{tcolorbox}
}

\definecolor{mylightgray}{RGB}{224,224,224}

\newboolean{showcomments}
\setboolean{showcomments}{true}
\ifthenelse{\boolean{showcomments}}
 { \newcommand{\mynote}[2]{
      \fbox{\bfseries\sffamily\scriptsize#1}
        {\small$\blacktriangleright$\textsf{\emph{#2}}$\blacktriangleleft$}}}
        { \newcommand{\mynote}[2]{}}

\newcommand{\projURL}{\url{https://github.com/Berickal/PEARL}}

\def\BibTeX{{\rm B\kern-.05em{\sc i\kern-.025em b}\kern-.08em
    T\kern-.1667em\lower.7ex\hbox{E}\kern-.125emX}}

\usepackage[accepted]{icml2025}

\usepackage{amsmath}
\usepackage{amssymb}
\usepackage{mathtools}
\usepackage{amsthm}

\usepackage[capitalize,noabbrev]{cleveref}

\theoremstyle{plain}

\theoremstyle{definition}

\theoremstyle{remark}

\usepackage[textsize=tiny]{todonotes}

\title{Memorization or Interpolation ? Detecting LLM Memorization through Input Perturbation Analysis}

\author{
  Albérick Euraste Djiré\\
  \texttt{euraste.djire@uni.lu}
  \and
  Abdoul Kader Kaboré\\
  \texttt{abdoulkader.kabore@uni.lu}
  \and
  Earl T. Barr\\
  \texttt{e.barr@ucl.ac.uk}
  \and
  Jacques Klein\\
  \texttt{jacques.klein@uni.lu}
  \and
  Tegawendé F. Bissyandé\\
  \texttt{tegawende.bissyande@uni.lu}
}
\date{}

\begin{document}

\maketitle

\begin{abstract}
While Large Language Models (LLMs) achieve remarkable performance through training on massive datasets, they can exhibit concerning behaviors such as verbatim reproduction of training data rather than true generalization. This memorization phenomenon raises significant concerns about data privacy, intellectual property rights, and the reliability of model evaluations.
This paper introduces PEARL, a novel approach for detecting memorization in LLMs. PEARL assesses how sensitive an LLM’s performance is to input perturbations, enabling memorization detection without requiring access to the model’s internals.
We investigate how input perturbations affect the consistency of outputs, enabling us to distinguish between true generalization and memorization. Our findings, following extensive experiments on the Pythia open model, provide a robust framework for identifying when the model simply regurgitates learned information. Applied on the GPT 4o models, the PEARL framework not only identified cases of memorization of classic texts from the Bible or common code from HumanEval but also demonstrated that it can provide supporting evidence that some data, such as from the New York Times news articles, were likely part of the training data of a given model.

\end{abstract}

\input{sections/intro}

\input{sections/background}
\input{sections/motivation}
\input{sections/approach}

\input{sections/setup}

\input{sections/experiments}

\input{sections/discussion}
\input{sections/relwork}
\input{sections/conclusion}

\balance
\bibliographystyle{icml2025}
\bibliography{main}

\input{sections/annexe}

\end{document}

%% file: sections/intro.tex
\section{Introduction}

The capabilities of LLMs are subject to ongoing debate and research within the AI community. In particular, while the reasoning performance of various models suggests that they excel at generalization, substantial empirical evidence exists that LLMs can occasionally \textit{memorize} specific examples from their training data, leading to outputs that reproduce verbatim content~\cite{carlini_extracting_nodate,hartmann_sok_2023}.

In the context of LLMs, memorization refers to the phenomenon where the model can reproduce or recall specific portions of text that it was exposed to during training. Let's consider a language model as a conditional probability distribution $p_\theta(y|x)$, where $\theta$ represents the model parameters, $x$ is the input context, and $y$ is the output token.

For a training dataset $D = {(x_i, y_i)}^N_{i=1}$, memorization can be defined through the concept of \textit{membership advantage}. This measures how differently a model behaves on inputs that were present in its training data versus those that were not. Formally, the membership advantage $ma$ for a model $M$ can be defined as:

\begin{center}
$ ma(M, x, y) = | p_\theta(y|x) - p_\theta(y|x'(x)) | $
\end{center}
Where $(x, y)$ is a sequence from the training data,  $x'(x)$ is a function that returns a similar but unseen sequence and $p_\theta(y|x)$ is the model's probability of generating $y$ given $x$.

A high membership advantage indicates that the model has memorized specific training examples rather than learning generalizable patterns since it assigns significantly different probabilities to seen versus unseen but similar examples.

Unfortunately, this mathematical modeling of memorization is actionable only when information about the training data is available. In practice, there is a lack of transparency around training data from major AI companies~\cite{mitStudyTransparency, europaActFirst}, due to several strategic and practical considerations: confidentiality represents a crucial competitive advantage in the industry and intertwines with significant legal risks. Additionally, detailed knowledge of training data could enable security exploits.

Yet, detecting memorization in LLMs, in particular commercial ones, is vital for multiple interconnected reasons. Without robust memorization detection, organizations risk privacy breaches, legal issues, and compromised model performance while potentially eroding public trust in AI systems. Indeed, from a privacy and security standpoint, memorization detection helps identify potential leaks of sensitive personal data \cite{yan_protecting_2024, carlini_extracting_nodate, lukas_analyzing_2023, yao_survey_2024}. 
From the perspective of assessing model quality, detecting memorization helps distinguish between true learning and mere regurgitation of training data, which is crucial for developing more reliable and generalizable AI systems. Legal compliance also benefits from memorization detection by helping prevent copyright infringement and managing intellectual property risks \cite{nytimes}. Perhaps most importantly, memorization identification supports transparency and trust in AI systems by enabling honest communication about model capabilities and limitations \cite{zhou_dont_2023}.

Early detection approaches analyzed model parameters to identify instances producing high-confidence generations~\cite{zhou_dont_2023}. However, as access to model parameters became increasingly restricted by proprietary LLMs, researchers are now investigating alternative black-box approaches. These methods include analyzing output distribution patterns across repeated prompts~\cite{zhou_quantifying_2024} and evaluating model responses using carefully crafted detection prompts~\cite{golchin_time_2024}.
While these approaches appear to be promising it remains an open question if they are successful in detecting actual memorization. Indeed, a critical challenge in memorization detection is distinguishing between two scenarios: \ding{182} \textit{True memorization}, where the model's high-confidence answers stem from reproducing training data and \ding{183}
\textit{Successful interpolation}, where the model demonstrates genuine generalization capabilities without relying on memorized content.

{\bf This paper.} In our study, we address the research question of how to identify true memorization in Black-box LLMs whose training data is further unknown. To that end, we propose the following {\bf Perturbation Sensitivity Hypothesis} (PSH):

\highlight{
For a given model, task, and data point, if the model has memorized that data point, then its task performance will exhibit high sensitivity to small input perturbations.
}

The remainder of the paper is organised according to the following contributions:

\begin{itemize}
    \item We introduce PSH in Section~\ref{sec:background}. After presenting background notions on memorization and interpolation, as well as related effort by prior work to estimate LLM's output distribution to identify memorization instances, we present our hypothesis and motivate how it correlates to memorization in Section \ref{sec:background}.
    \item We propose PEARL (PErturbation Analysis for
Revealing Language model Memorization) a framework that builds on PSH, to predict memorization in LLMs in Section \ref{sec:methodology}.
    \item  We empirically demonstrate the power of PSH by evaluating its effectiveness on the open-source model Pythia with its associated training dataset Pile in  Section \ref{sec:exp}. 
    \item We use the Bible, NY Times and HumanEval datasets to develop case studies of memorization within GPT-4o in Section \ref{sec:exp}.
\end{itemize}

\begin{figure*}[!ht]
    \centering
    \includegraphics[width=1\linewidth]{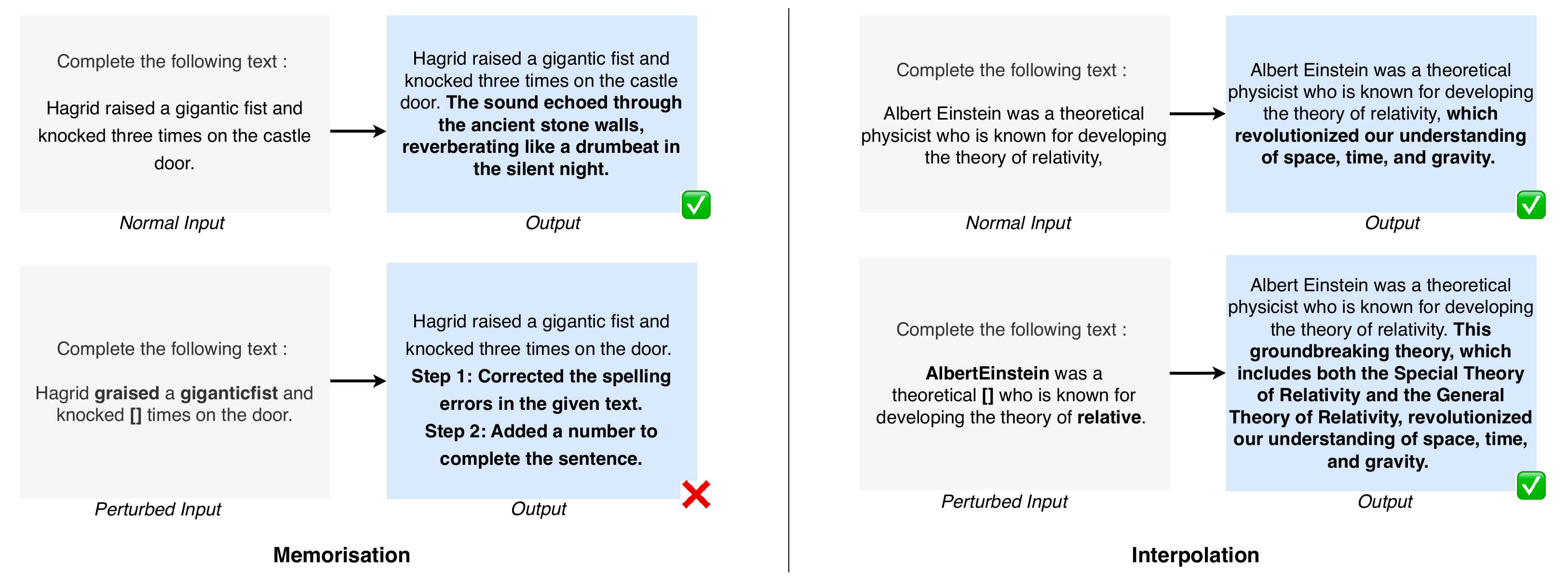}
    \vspace{-4mm}
    \caption{Illustration of memorisation and interpolation in a Completion task with the model amazon-nova-lite-v1.0 \cite{intelligence2024amazon}}
    \label{fig:memorization_illustration}
\end{figure*}

%% file: sections/background.tex
\section{PSH: Perturbation Sensitivity Hypothesis} \label{sec:background}

LLMs operate as probabilistic models that aim to learn and generalize patterns from their training data. While generalization is a key objective, model reliability is intrinsically tied to the quality and characteristics of data used during pretraining, fine-tuning, and reinforcement learning. This creates an inherent tension: models must balance between generalizing from learned patterns and maintaining fidelity to their training data to ensure reliable outputs.

The concept of memorization in LLMs has multiple formal definitions in the literature. One prominent framework focuses on verbatim reproduction, where a model outputs exact sequences present in its training data~\cite{carlini_extracting_nodate, yao_survey_2024}. This phenomenon can manifest across various tasks, including sequence completion, question-answering, and reasoning~\cite{hartmann_sok_2023}. While memorization is often associated with overfitting, particularly during fine-tuning when models are repeatedly exposed to the same data~\cite{schwarzschild_rethinking_2024, duan_membership_2024}, recent work by~\cite{dankers_generalisation_2024} suggests that memorization may be a necessary precursor to generalization. Their findings indicate that LLMs exhibit heightened sensitivity to memorized outliers compared to more common data patterns.
In contrast, interpolation refers to a model's ability to identify and leverage underlying patterns in the training data to generate novel outputs. More formally, given input $x$ from the input space $\mathcal{X}$, an interpolating model produces outputs by computing the expected value over its learned distribution, thus reflecting learned patterns while maintaining some measure of uncertainty or confidence in its predictions. This can be expressed as:
\begin{equation}
f(x) = \mathbb{E}_{y \sim p(y|x)}[y]
\end{equation}
where $f(x)$ is the model's output, and the expected value, $\mathbb{E}[y]$, is taken over outputs $y$ sampled from the learned conditional probability distribution $p(y|x)$.

Figure~\ref{fig:memorization_illustration} illustrates real-world examples of memorization and interpolation that we have witnessed using the amazon-nova-lite-v1.0 model. We consider two text samples, one from the Harry Potter series and the other from Wikipedia. When the sample data are submitted as such as inputs with a prompt for text completion tasks, the model produces correct outputs. The question is then : ``\textit{how do we know whether the model is interpolating well or whether it has memorized the result from its training data}''. 

Building on the PSH, we introduce some perturbations to the inputs and prompt again the model for completion. The model clearly fails for the Harry Potter text. In contrast, for the Wikipedia text about Albert Einstein, the model can complete the task.

\begin{figure}[!ht]
     \centering
     \vspace{-0.45cm}
     \includegraphics[width=1\linewidth]{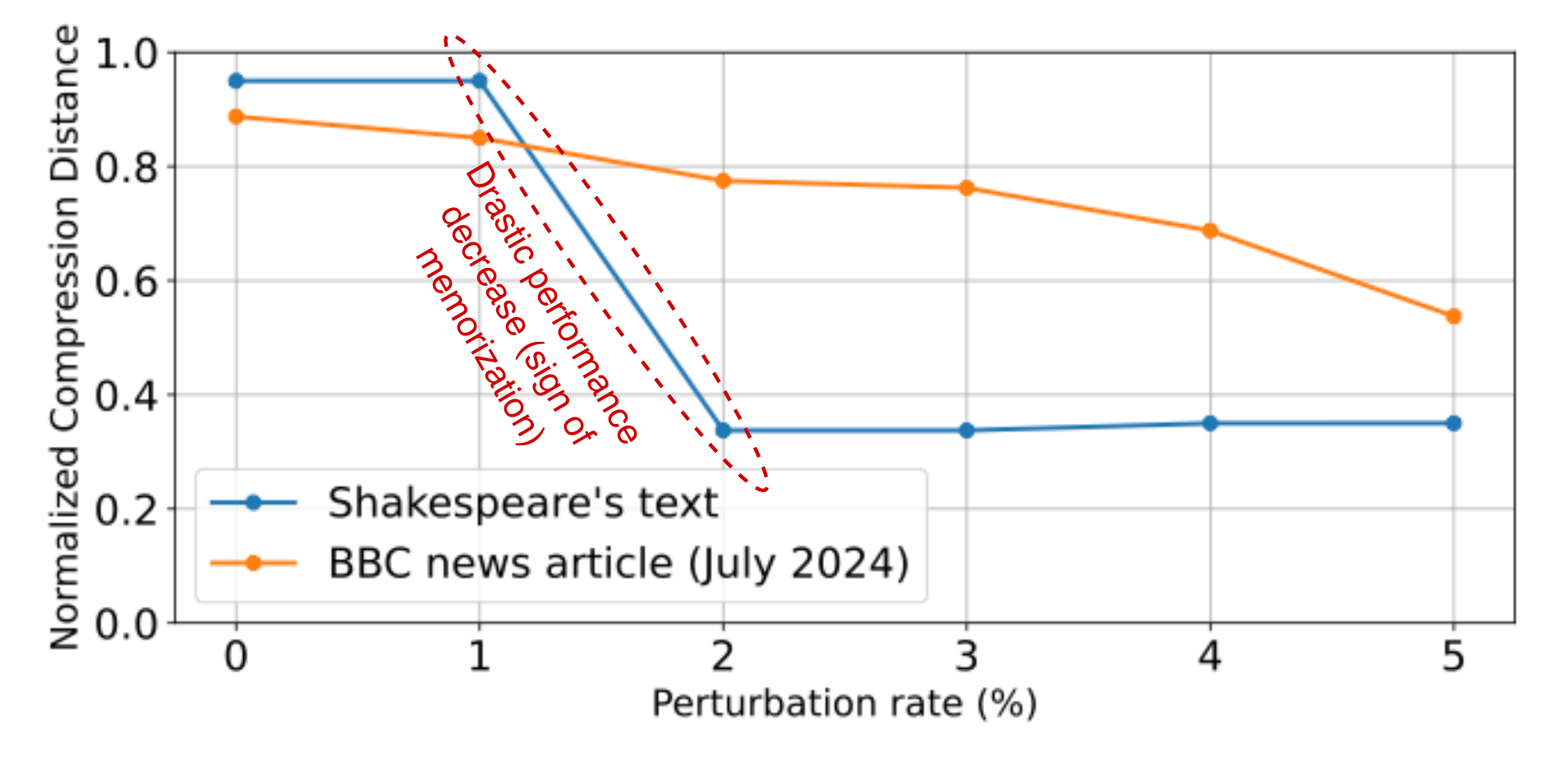}
     \caption{ GPT\_4o text completion \textbf{performance falloff }for a memorized Shakespeare poem submitted to perturbations vs regular performance degradation with a recent text (not part of the training set of GPT\_4o).}
     \label{fig:motivation_example}
 \end{figure}

Generalizing on these examples, we postulate that in case of memorization, the LLM output is sensitive to the training data. Thus, a certain degree of perturbation to a memorized input data can lead to a drastic drop of performance from the model. 
Consider Figure~\ref{fig:motivation_example}: we consider the GPT-4o model for a text completion task using data text samples from Shakespeare poem (confirmed by GPT-4.0 as part of its training data) and a news article produced after the release of GPT-4o provided in the annex \ref{annexe:1}. As perturbation, we consider random bit flips of the inputs (cf. Section~\ref{sec:methodology}). By measuring the performance (in terms of Normal Compression Distance) of the model for this task, we note that for the first sample text, the model performance drops abruptly when we apply beyond a given perturbation rate (2\%).  In contrast, the performance decrease is more regular for the second sample text (which we know could not have been memorized).
This example evidences the PSH hypothesis for identifying memorization.

%% file: sections/approach.tex
\section{PEARL: PErturbation Analysis for Revealing Language model Memorization} \label{sec:methodology}

We design PEARL as a novel framework that builds upon our PSH hypothesis, which posits that memorized data points exhibit high sensitivity to small input perturbations in terms of task performance. By systematically implementing this hypothesis through controlled input variations, PEARL provides a robust method to detect memorization in LLMS. 

\begin{figure*}
    \centering
    \includegraphics[width=1\linewidth]{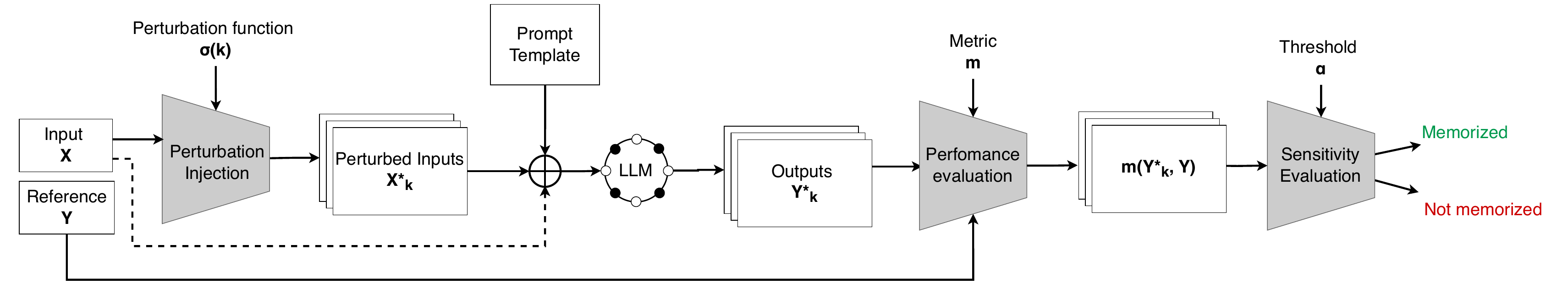}
    \caption{Overview of the PEARL framework for identifying memorization in LLMs based on the PSH hypothesis}
    \label{fig:workflow}
    \vspace{-0.5cm}
\end{figure*}

As illustrated in Figure~\ref{fig:workflow}, the framework develops a comprehensive analytical pipeline that quantifies perturbation sensitivity patterns associated with performance drops (based on applied thresholds), enabling reliable distinction between memorized content and genuinely learned patterns. This approach offers researchers and practitioners a principled way to assess memorization in language models without requiring detailed information about neither internal parameters of a model nor its training data.

Given an input $X$, a reference output ($Y$) can be known: in the text completion task, the input is a subsequence of a sample text which is therefore reference output. For other tasks where the output is not directly available in the sample texts, the reference output is obtained by applying the model to the input without any perturbations. A perturbation is then applied to the input before it is submitted to the LLM. PEARL then computes the difference of performance that the LLM achieves between the reference output and the output associated with the perturbed input. This process is repeated several times using different intensities of perturbations. Then PEARL quantifies the sensitivity of the model to those perturbations to decide whether it considers the original input data as being memorized or not. We detail in the remainder of this section the different steps of the PEARL approach.

\subsection{Input Perturbations Generation}
\begin{figure}[!hhh]
    \centering
     \vspace{-0.5cm}
  \includegraphics[width=0.8\linewidth]{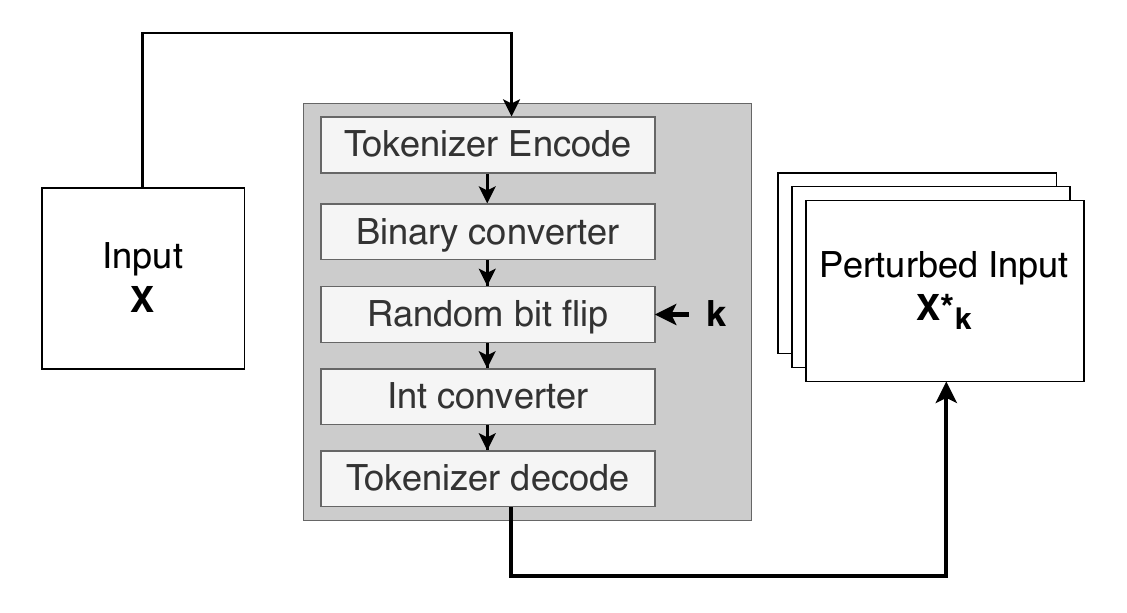}
    \caption{Bit-flip perturbation injection in text inputs, with \(k \in \mathbb{N}\) }
    \label{fig:xp_illustration}
     \vspace{-0.5cm}
\end{figure}

The first step of PEARL involves generating controlled perturbations to test inputs. We consider a perturbation function $\sigma(k)$ that systematically modifies input data with a given intensity $k$. We propose in this work to focus on a single perturbation function for all text inputs and for all tasks: Figure~\ref{fig:xp_illustration} details the process for the bit-flip perturbation generation. The function thus consists in flipping $k$ bits in the binary version of the input text and then decoding it back into text to yield the perturbed version of the input. 

\subsection{Prompting the LLM for specific tasks}

Given a task $T$, an input $X$ and the set of generated perturbed inputs \(X^* = \{x^*_0, x^*_1, ... x^*_n\}\), we prompt the LLM and collect a set of output sets \(Y^* = \{Y^*_0, Y^*_1, ... Y^*_n\}\) with \(Y^*_k = \{y^*_{(k, 1)}, y^*_{(k, 2)}, …. y^*_{(k, i)} \}\) being the set of outputs associated with the perturbed input \(x^*_k\). 
Indeed, for each input, we prompt the LLM $i$ times to obtain $i$ sample outputs. This is done in order to help establish statistical significance of results as single runs might be outliers and not representative of true model capabilities. 

\subsection{Quantifying LLM Sensitivity to Input Perturbation}
For a given output set \(Y^*_k\) (yielded by an LLM prompted with perturbed input \(x^*_k\)) and a reference output \(Y\), we compute a metric on the performance variation as follows:
\begin{equation} \label{eq:2}
    m(Y^*_k) = \frac{1}{i} \sum_{j=1}^idistance(y^*_{(k,j)}, Y) 
\end{equation}
where $distance$ implements a task-dependent edit distance function.  

To quantify the model's sensitivity to the perturbations on input $X$, we compute the maximum performance falloff among consecutive perturbation intensities:
\begin{equation} \label{eq:3}
    sensitivity(X) = {\underset{j \in {1,..,k-1}}{max}(m(Y^*_j) - m(Y^*_{j+1})) }
\end{equation}

\subsection{Deciding on Memorization}
Input $X$ is then identified as a memorized instance of the LLM training data based on the following condition :
\[
memo(X) = 
    \begin{cases}
         sensitivity(X) > \alpha\ & \text{memorized} \\
        otherwise & \text{not memorized}
    \end{cases}
\]
where \(\alpha\) represents a hyperparameter threshold that defines the \textit{significance of the task performance falloff}.

%% file: sections/setup.tex
\subsection{Experimental Setup} \label{sec:setup}
We evaluate PEARL to validate PSH through two series of experiments. The first one considers open-source models with transparent details about the training data. We even perform fine-tuning on our true positive test set, with multiple epochs, to ensure that the data that will be tested for memorization has indeed been considered by the model.
The second experiment considers a close model for developing case studies in the wild, toward demonstrating the potential application of PEARL for membership inference.

\paragraph{Model Selection.} We consider Pythia~\cite{biderman_pythia_2023}, a suite of open-source models designed to facilitate scientific research. It includes a diverse range of model sizes, spanning from smaller configurations, such as 70M parameters, to larger configurations with billions of parameters, including 6.9B and 13B. Additionally, for each model size version, two variants are provided: one is trained on the full pretraining dataset, \textit{The Pile}, while the other is trained on a deduplicated version of \textit{The Pile}. 

In this study, we focused on the variant of Pythia models that were trained on the deduplicated dataset. Indeed, eliminating duplicate content from the training dataset minimizes the likelihood of memorization arising from repeated exposure to identical data. Thus, this setup enable us to systematically evaluate memorization under controlled conditions, ensuring that any identified instances of memorization can be attributed to single occurrences of data rather than repeated patterns.

To validate our approach in real-world scenarios, we extended our analysis to GPT-4o, examining memorization detection in a black-box context where model internals parameters are inaccessible. The selection of GPT-4o is particularly relevant given recent developments in AI ethics and intellectual property rights. Notably, in light of the January 2024 legal dispute between OpenAI and The New York Times regarding alleged copyright infringement and unauthorized use of journalistic content for training purposes \cite{nytimes}. We thus proposed to investigate potential memorization of New York Times articles in GPT\_4o.

\paragraph{Datasets.}
Pythia models are pre-trained on the Pile dataset~\cite{gao_pile_2020}, a diverse corpus curated to encompass a broad spectrum of text types, ranging from literature and scientific articles to web content and programming code. From an experimental perspective, we posit that any data sample in this dataset could have been memorized by the model. We select a random subset of 1\,000 text samples, each having at least 300 tokens, to build our positive set (memorization is ``forced'' somehow by multiple rounds of fine-tuning to drive overfitting). Conversely, we can assert that data absent from \textit{the Pile} could not have been memorized, and the model’s high performance on such data would likely be due to its generalization capabilities. We consider for negative samples (non-memorized data) the RefinedWeb dataset, which includes a collection of texts that is disjoint from \textit{the Pile}, providing a reliable benchmark for evaluating the model’s generalization capabilities on unseen~\cite{penedo_refinedweb_2023}. The set selection is the same as Pythia: 1\,000 random samples.

For experiments in the wild, we build experimental sets of 100 data samples collected from four different sources: 

\ding{182} The \textit{HumanEval Dataset}, released by OpenAI, contains samples for 164 Python programming problems. Each problem includes a function signature, the documentation (\textit{docstring}), the implementation body, and multiple unit tests. This dataset serves as a standard benchmark for assessing code generation capabilities in language models \cite{chen2021evaluating}. Since it is publicly available, it is suspected to have been memorized by GPT-4o~\cite{matton_leakage_2024} or at least is part of its training data~\cite{dong_generalization_2024}.

\ding{183} The \textit{Less Basic Python Programming (LBPP)} is a programming challenge dataset published on July 11th, 2024, subsequent to GPT-4o's release. Similar to HumanEval, it focuses on Python programming tasks but emphasizes more complex programming concepts and algorithmic challenges, and has explicitly been proposed~\cite{matton_leakage_2024} to address the contamination of models, such as GPT-4o, by existing code generation test sets, such as HumanEval. 

\ding{184} The \textit{Bible} is an ancient, popular and unique collection of religious texts. Due to its open nature, the Bible has been used as one of the key source for training data of most models, alongside Wikipedia. Due to its unique writing style among other more contemporary text samples, we assume that Bible verses are more likely to be memorized. 

\ding{185 } The \textit{New York Times (NYT)} is a newspaper that publishes articles daily on various topics including politics, business, technology and culture. The use of NYT articles by OpenAI is the subject of a lawsuit~\cite{nytimes} for copyright infringement. The argument of OpenAI refer mostly to the  ``fair use'' rules\footnote{\scriptsize It is a doctrine in US law that allows copyrighted material to be used for educational, research or commentary purposes. In order to clear the fair use test, the work in question has to have transformed the copyrighted work into something new, and the new work cannot compete with the original in the same marketplace, among other factors.}. In this paper, we will consider that NYT data is known to be part of the training set of GPT-4o. Therefore, if evidence can be provided that some samples have been memorized, the fair use defense can be questioned.

\paragraph{Tasks \& Evaluation Metrics.}
We consider four common tasks in Natural language processing and Software engineering: text completion and text summarization as well code completion and code summarization. We distinguish the tasks because code summarization the input is code written in a programming language and the output is text written in natural language. 

We use different metrics for model performance evaluation depending on the tasks. For code/text completion, we rely on the Normalized Compression Distance (NCD). For code/text summarization, we use ROUGE-L~\cite{lin-2004-rouge}.

{\bf Hyperparameters.}
Table \ref{tab:hyper_parameters_list} summarises the different hyperparameter values used in our experiments.
\begin{table}[!t]
    \centering
    \scriptsize
    \begin{tabular}{p{.5cm}|p{3.5cm}|p{2.5cm}}
        Param. &  Description & Value \\
        \hline
        $k$ & Hyperparameter that allow to control the perturbation intensity. In the case of bitflip, it consisted in the percentage of token modified in the input. (cf.  Section~\ref{sec:controled_env}) & \(k \in \{0, 1, 2, 3, 4, 5\} \) \\
        \hline
        $i$ & The number of generated output per perturbed input. (cf. Section~\ref{sec:controled_env}) & \(i = 10\) \\
        \hline
        \(\alpha\) & Threshold that allow to determine whether an instance is memorized or not  (cf. Section~\ref{sec:exp}) & Depend on the model, task and dataset considered.  \\
    \end{tabular}
    \caption{Hyperparameters values}
    \label{tab:hyper_parameters_list}
         \vspace{-0.5cm}
\end{table}

%% file: sections/experiments.tex
\section{Experimental Results}\label{sec:exp}

\subsection{Research questions}

\begin{enumerate}
\item \textbf{[RQ1.] \textit{Can the input Perturbation Sensitivity Hypothesis provide a reliable indicator for detecting and quantifying memorization in LLMs?}} Most prior works in the literature have assessed memorization as a membership inference problem. In the absence of ground truth, we propose a 2-step validation approach in a controlled environment (open model and data): first, we assess the discriminative power of PEARL to suspect memorization within a dataset that is not part of the training data and dataset that is part of the training dataset. Then, we evaluate whether the number of memorized instances identified by PEARL evolves consistently with the fine-tuning effort to achieve overfitting (hence memorization). 

\item \textbf{[RQ2.] \textit{In which specific datasets does PEARL find memorization instances when applied to GPT-4o?}} We develop cases studies with a closed-source model investigating memorization decisions for a variety of datasets - including datasets that are known to be part of the training datasets, that are suspected by prior work to have been memorized and that is known to not be part of the training datasets.

\item \textbf{[RQ3.] \textit{To what extent the memorization prediction is task-dependent in PEARL?}} In this research question, we investigate whether the discriminative power of PEARL is stable during the analysis of the performance falloffs across the two considered tasks.

\end{enumerate}

\subsection{Validation of PSH} \label{sec:controled_env}
\begin{figure}[!h]
    \centering
    \includegraphics[width=0.7\linewidth]{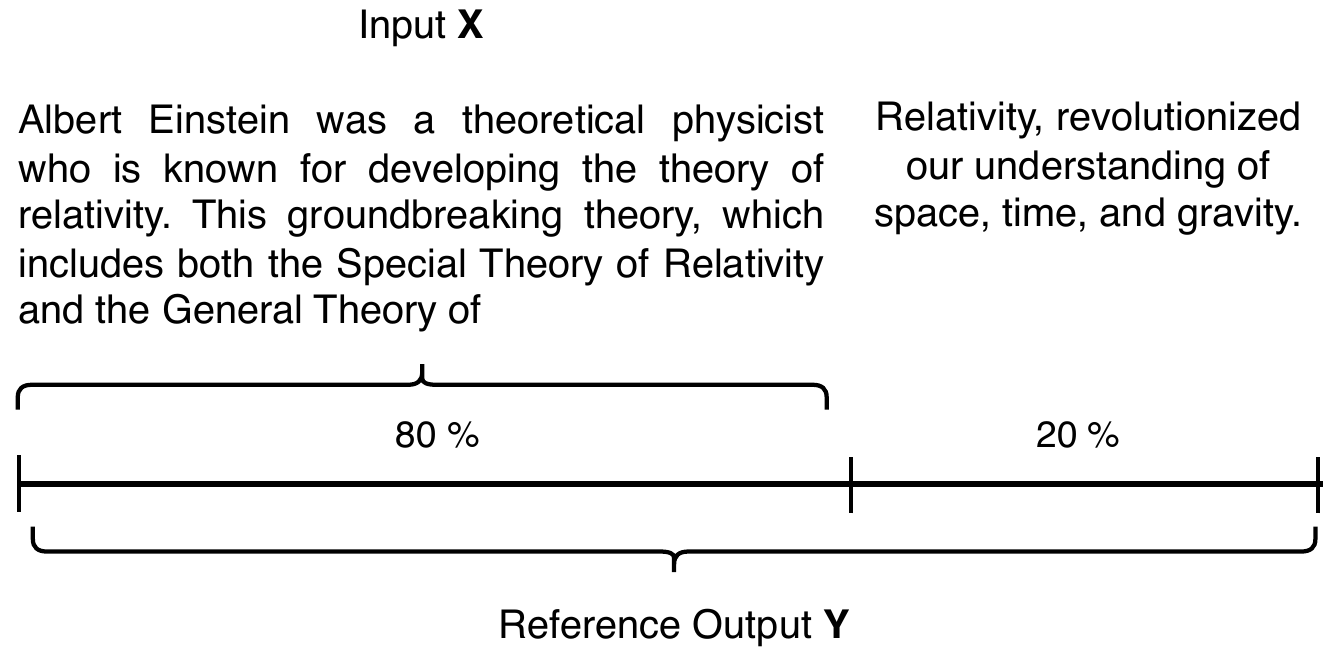}
    \vspace{-0.2cm}
    \caption{Examples input \textit{X} and reference output \textit{Y} for a text completion task.}
    \label{fig:text-split}
    \vspace{-0.2cm}
\end{figure}

Pythia being a generative model which was not trained to follow instructions, we apply its 410m version only for text completion tasks.
Figure~\ref{fig:text-split} illustrates how each sample text is split to derive the input to be used in the prompt, as well as the reference output. After generated the perturbed inputs and applying the repetitive prompts, we collect the outputs and compute the performance metrics (i.e., NCD) and the model sensitivity metric as defined in Equations~\ref{eq:2} and~\ref{eq:3}.

\begin{figure}[!h]
    \centering
    \includegraphics[width=0.7\linewidth]{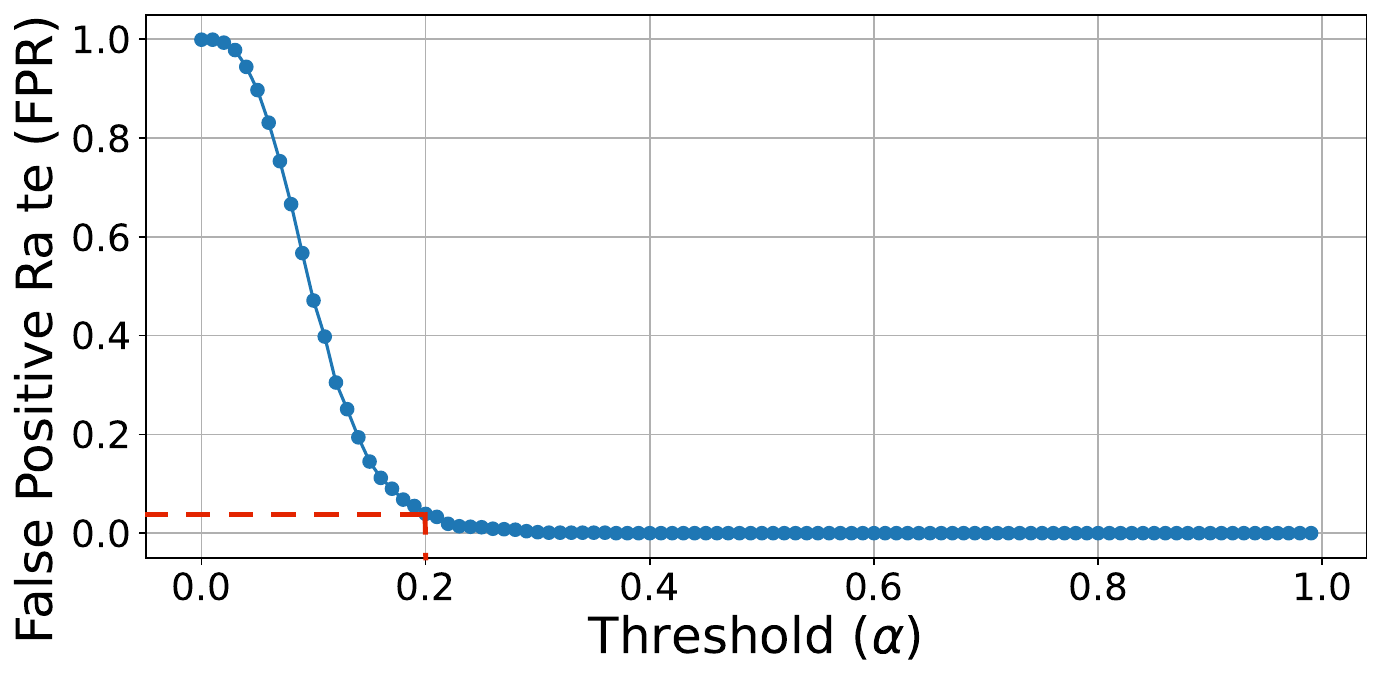}
    \caption{Evolution of FPR of PEARL following the variations of sensitivity threshold \(\alpha\) when attempting to identify memorized instances in a dataset that is known to not be part of the training set of the Pythia model}
    \label{fig:threshold_impact}
\end{figure}

We consider the negative set (i.e., RefineWeb, which is not part of the training set). Figure~\ref{fig:threshold_impact} represents the different false positive rates of PEARL on this dataset when the sensitivity threshold \(\alpha\) is varied. Given that a threshold \(\alpha = 0.2\) leads to a low false positive rate (0.04), we set it as the threshold value for the experiments that are conducted on this task and with this model.

\begin{figure*}[t]
    \centering
    \begin{minipage}[t]{0.48\linewidth}
        \centering
    \includegraphics[width=0.7\linewidth]{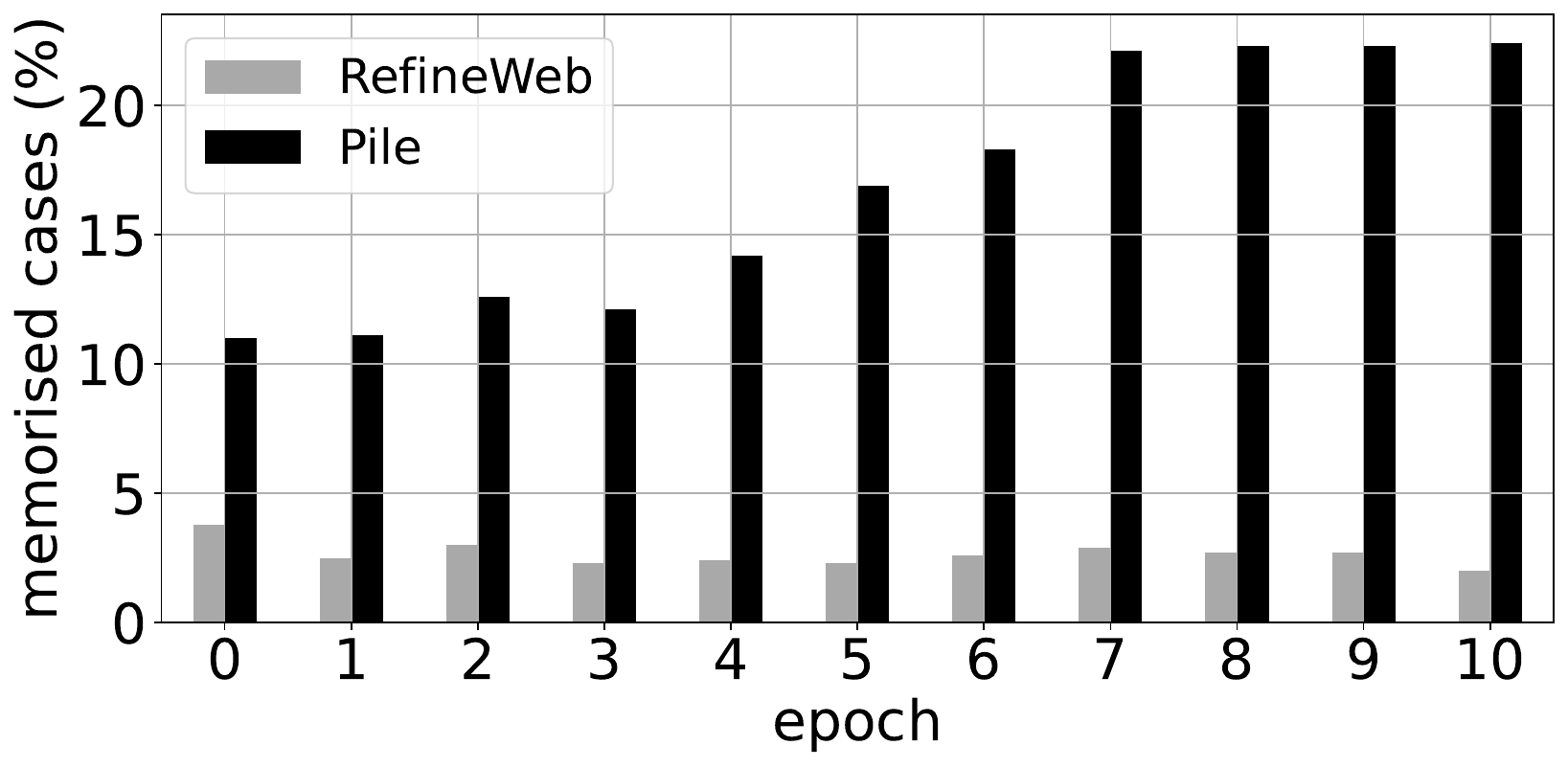}
    \vspace{-0.3cm}
    \caption{Proportion of data samples identified the RefineWeb and the Pile datasets as being memorized by Pythia-410m - \(\alpha = 0.2\)}
    \label{fig:result_epoch}
    \end{minipage}
    \hfill
    \begin{minipage}[t]{0.48\linewidth}
        \centering
    \includegraphics[width=0.7\linewidth]{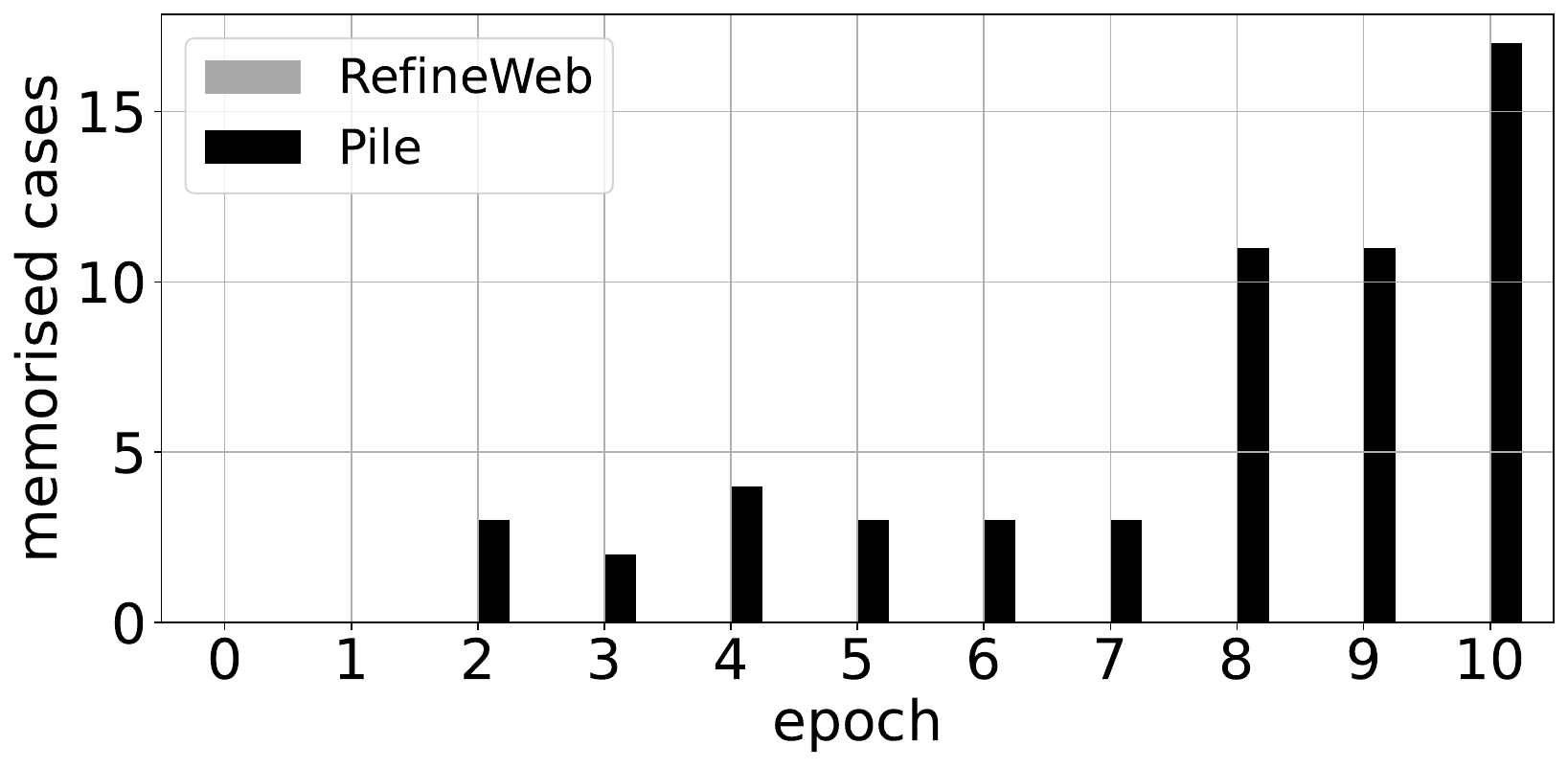}
       \vspace{-0.3cm}
 \caption{Number of instances identified in the RefineWeb and the Pile datasets as being memorized by Pythia-410m  - \(\alpha = 0.4\)}
    \label{fig:high-alpha}
    \end{minipage}
    \label{fig:combined}
\end{figure*}

Applying PEARL with the set sensitivity threshold \(\alpha=0.2\), we compare the proportion of PEARL-identified memorized instances between the RefineWeb dataset (not part of the training dataset) and the Pile dataset (training dataset). In Figure~\ref{fig:result_epoch}, we note that PEARL identifies an order of magnitude more samples of the Pile ($> 20\%$ at epoch \#10) as memorized compared with the number of samples from RefinedWeb ($\simeq2\%$ at epoch \#10). With a higher sensitivity threshold (e.g., \(\alpha=0.4\) in Figure~\ref{fig:high-alpha}) we can eliminate the number of false positives for RefineWeb, albeit a stricter identification cutoff for the Pile. 

The results further show that as we iterate in the fine-tuning process, the number of detected memorized instances increases for the Pile (training data) while the number of detected instances remains stable and low for RefineWeb (not part of the training data).

\highlight{
\footnotesize
{\bf Answer to RQ1:} PSH proves to be a reliable framework for detecting memorization in LLMs as evidenced by the clear distinction in perturbation sensitivity between training data (Pile) and non-training data (RefineWeb) from the Pythia model, with adjustable detection thresholds enabling fine-grained control over false positive rates. The increasing detection rate in training data across fine-tuning epochs, coupled with stable low rates in non-training data, further validates PSH as a robust indicator of memorization. These findings establish PSH as a principled foundation for memorization detection in LLMs, with PEARL offering a practical implementation. }

Finally we compare our detection results against the detection results obtained with a whitebox approach (Adversarial Compression Ratio~\cite{schwarzschild_rethinking_2024}) on 100 samples of the FamousQuotes dataset. Results show that our detections significantly overlap: the advantage of PEARL, being that it does not require knowledge on the models internals, nor of the training data. Detailed results are provided in Annex~\ref{comp:agc}.

\subsection{Impact of Model Size}
Prior work from Google Research~\cite{carlini_quantifying_2023} has argued that larger models have increased capacity for memorization. 
We repeat the experiments of RQ1 with different size-variants of Pythia and confirm this finding: in Table~\ref{tab:memorised_model_size}, the number of identified memorized instances increases with the model size, even for high sensitivity thresholds.

\begin{table}[!h]
    \centering
    \scriptsize
    \begin{tabular}{l|p{1.6cm}|p{1.6cm}|p{1.6cm}}
        Model & \#identified memorized instances \linebreak \(\alpha = 0.16\) & \#identified memorized instances  \linebreak \(\alpha = 0.2\) & \#identified memorized instances \linebreak \(\alpha = 0.25\) \\
        \hline
        Pythia 410m & 265 & 110 & 32 \\
        \hline
        Pythia 1.4B & 349 & 171 & 57 \\
        \hline
        Pythia 6.9B & \textbf{449} & \textbf{259} & \textbf{110}  \\

    \end{tabular}
    \caption{Number of memorization cases identified by PEARL on Pythia models of different size}
    \label{tab:memorised_model_size}
\end{table}


\subsection{GPT-4o Memorizations} \label{sec:real_case}
We consider the sample sets from the following sources, which have varying probability of being memorized. 

\begin{figure}[!h]
    \centering
    \includegraphics[width=0.8\linewidth]{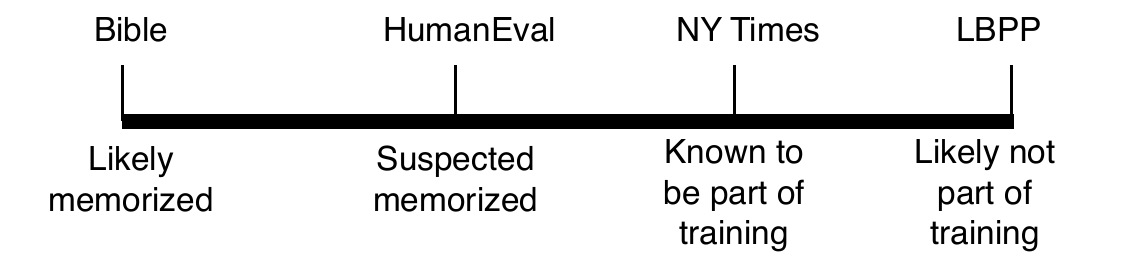}
  
\end{figure}
\vspace{-4mm}

We apply code or text completion tasks depending on the dataset. Given that LBPP is constituted of samples that are less likely to not be part of the training set of GPT-4o according to Matton et al. \cite{matton_leakage_2024}, we apply first PEARL on it to select the sensitivity threshold values that minimize the number of false positives (cf. Annex~\ref{annexe:alpha_gpt_4o_completion} for the graph on the evolution of FPR vs sensitivity threshold $\alpha$ variations).

Table~\ref{tab:gpt_4_sources} reports on the number of memorization cases that PEARL identifies among each subset of 100 samples within the Bible, HumanEval and NY Times datasets. 

\highlight{\footnotesize {\bf Answer to RQ2:} PEARL's analysis of GPT-4o reveals a non-uniform pattern of memorization across datasets, with significant memorization detected in HumanEval, moderate levels in the Bible, and fewer instances in NY Times content. The results suggest that memorization is more likely to occur in unique or stylistically distinctive content rather than uniformly across training data.}

We provide in Annex~\ref{annexe:low_high_sensitivity}\ examples cases where the computed sensitivity values are the lowest and the highest in the Bible and the NYT datasets, suggesting that memorized samples are outliers in the dataset (due to style (e.g., for the Bible) and/or uniqueness of the content (e.g., for the NYT)). 

\begin{table}[]
    \centering
    \scriptsize
    \begin{tabular}{l|p{1.6cm}|p{1.6cm}|p{1.6cm}}
         Data source & \#identified memorized instances \linebreak \(\alpha = 0.14\) &  \#identified memorized instances \linebreak \(\alpha = 0.24\) &  \#identified memorized instances \linebreak \(\alpha = 0.29\) \\
         \hline
         HumanEval & \textbf{60} & \textbf{30} & \textbf{23} \\
         \hline
         Bible & 42 & 3 & 2 \\
         \hline
         NY Times & 5 & 1 & 0 \\
    \end{tabular}
    \caption{Number of memorization cases identified by PEARL in the GPT-4o model. Example cases are presented in Annex~\ref{annexe:2}}
    \vspace{-0.7cm}
    \label{tab:gpt_4_sources}
\end{table}

\subsection{Task Dependence of PSH} 
We investigate to what extent the identification of memorization based on PSH is influenced by the choice of the task. To that end, we run again all experiments on GPT-4o with the tasks of code/text summarization for the same datasets (Bible, HumanEval, NYT).  
We recall that this experiment is not possible with the Pythia model which does not support summarization tasks.
We use again the LBPP dataset to calibrate the values of $\alpha$ so as to minimize the false positive rates of PEARL (cf. Annex~\ref{fig:lim_summary_gpt_4}).

\begin{figure}[!h]
    \centering
    \hspace*{-0.4cm}
    \includegraphics[width=0.8\linewidth]{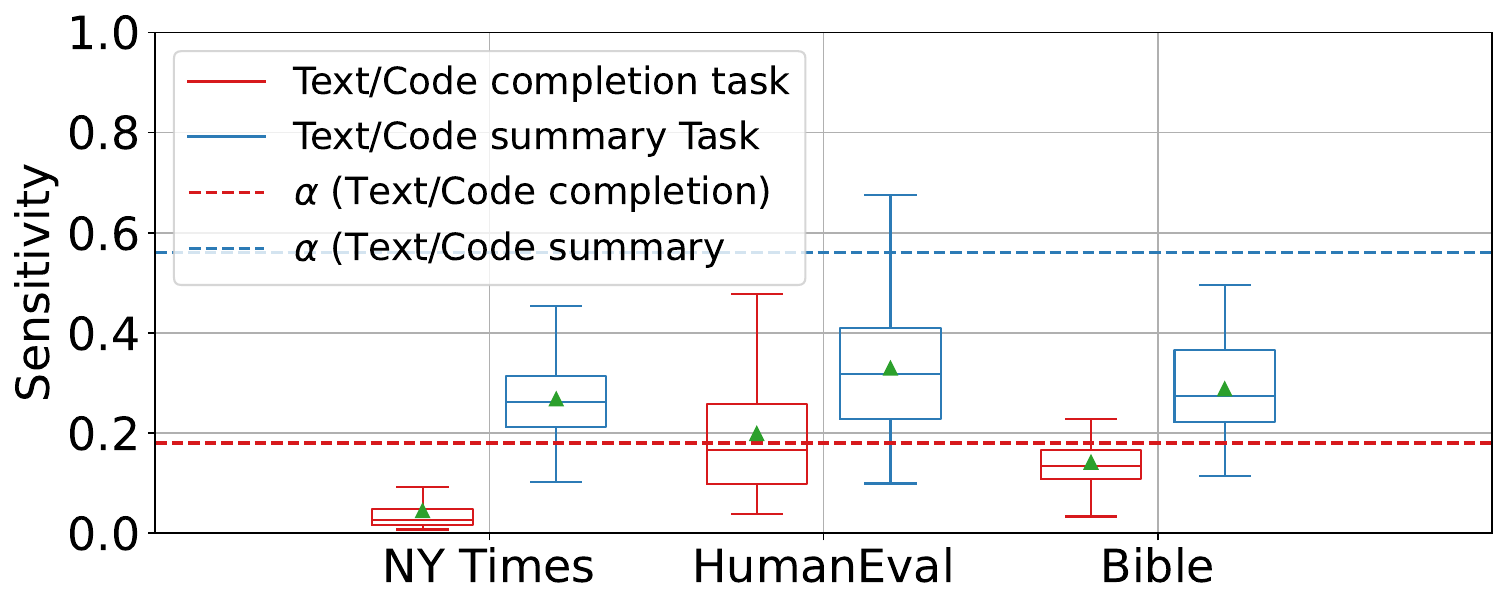}
       \vspace{-0.3cm}
 \caption{Distribution of GPT-4o sensitivity measurements to input perturbations for different data sources}
    \label{fig:sensitivity_distribution}
\end{figure}

The experimental results show that among all samples identified previously  as memorized by PEARL (using the completion task), only 6 are still identified when using summarization tasks. Figure~\ref{fig:sensitivity_distribution} shows the distribution of sensitivity values computed for each sample based on the two tasks: with the completion tasks, we can identify memorization cases of GPT-4o with a large number of HumanEval code samples and some Bible verses. There is also a statistically significant difference between the medians of sensitivity values for these Human Eval and NYT.  In contrast, with the summarization tasks, only a few samples from HumanEval are identified as memorized. The difference between Human Eval and NYT is further not significant.


This contrast can be explained by the fact that, in most cases, the context provided by the input is sufficient for GPT-4o to perform summarisation tasks. Hence, the perturbations introduced in the inputs do not lead to drastic performance falloffs in many cases.

\highlight{\footnotesize {\bf Answer to RQ3:} PSH's effectiveness for memorization detection varies significantly with task type. While PEARL reveals clear memorization patterns in completion tasks, these patterns largely vanish in summarization tasks, suggesting that perturbation sensitivity better exposes memorization when the model needs to reproduce exact content rather than generate new content based on understanding.
}

%% file: sections/discussion.tex
\section{Discussion}
\subsection{Impact}
Our validation of PSH through PEARL provides the AI community with a principled framework for detecting memorization in LLMs, towards understanding and quantifying how models store and use training data. For practitioners, PEARL offers a practical tool to assess memorization risks, while for researchers, our findings open new avenues for investigating the relationship between memorization, task types, and model performance. 

Memorization appears inevitable for unique content (like Shakespeare or Bible verses) where models lack similar examples for learning general patterns. By providing a reliable method to detect such cases, our research contributes concrete evidence to ongoing debates about data ownership and consent in AI training.

\subsection{Limitation}
PEARL's sensitivity threshold ($\alpha$) calibration requires access to data known to be outside the model's training set. Thus, such a calibration is challenging for closed-source models where training data information is unavailable.

PEARL's effectiveness varies across task types, as shown by the contrasting results between completion and summarization tasks. This task dependency indicates that perturbation sensitivity may not be equally revealing of memorization across all model capabilities.

\subsection{Threats of validity}
Regarding external validity, our findings from GPT-4 and Pythia models may not generalize to other architectures, and our selected datasets may not represent all content types. For internal validity, our calibration assumes LBPP data is truly outside the training set, while our perturbation techniques could introduce unintended biases. For construct validity, by relying on distance metrics to measure differences between outputs from original and perturbed inputs, we may not capture all semantic aspects of model responses. Furthermore, our binary classification of content as either memorized or non-memorized likely oversimplifies what is  a spectrum of memorization behaviors in LLMs. This dichotomous approach, while practical for detection purposes, may not fully represent the nuanced ways in which models incorporate and utilize training data. However, these limitations are partially mitigated by PEARL's configurable sensitivity threshold, which allows practitioners to adjust the detection granularity based on their specific needs. Additionally, our demonstration of consistent patterns across different datasets and models suggests that despite these simplifications, the approach successfully captures meaningful memorization signals.

%% file: sections/relwork.tex
\section{Related Work}

Research on memorization in LLMs has evolved along several directions. Early studies focused on extracting sensitive information through Membership Inference Attacks~\cite{duan_membership_2024}. While pioneering, they  required access to model internals such as loss functions and architecture details, limiting their applicability to black-box scenarios.

A different perspective emerged with Schwarzschild et al.'s~\cite{schwarzschild_rethinking_2024} definition of memorization based on Adversarial Compression Ratio (ACR), where content is considered memorized if reproducible through significantly shorter prompts. This approach leverages GCG~\cite{zou2023universal} for finding minimal inputs that generate target outputs, providing a novel metric for memorization detection when model generation parameters are accessible. 

Recent work has deepened our understanding of memorization mechanisms. Speicher et al.~\cite{speicher_understanding_2024} characterized learning phases in memorization, while Dankers et al.~\cite{dankers_generalisation_2024} demonstrated that memorization occurs gradually across model layers, with early layers playing a more crucial role. Their work also provided evidence for the task-dependency of memorization, aligning with our findings.

%% file: sections/conclusion.tex
\vspace{-0.3cm}
\section*{Conclusion}
This paper introduces PEARL, 
a framework that validates and operationalizes the Perturbation Sensitivity Hypothesis (PSH) for detecting memorization in LLMs. Through experimentation with GPT-4 and Pythia models across diverse datasets, we demonstrate that perturbation sensitivity serves as a reliable indicator of memorization, particularly in tasks requiring exact content reproduction. Our findings reveal that memorization patterns are strongly task-dependent and more prevalent in unique or distinctive content. As AI systems continue to scale, PEARL offers valuable insights for addressing data privacy concerns as well as biases in model evaluation and contributes to ongoing discussions about responsible AI development and training data usage.

%% file: sections/annexe.tex

\newpage

\section*{Annexes}
\subsection*{DATA AVAILABILITY}
For the sake of Open Science, we provide to the community all the
artifacts used in our study. The project’s repository including all
artifacts (tool, datasets, etc.) is available at:

\vspace{1mm}
\begin{center}
    \projURL
\end{center}

\label{comp:agc}

\subsection*{Annex 1 : Adversarial Compression Ratio (ACR) vs Perturbation Sensitivity Hypothesis (PSH)}
\begin{figure}[!h]
    \centering
    \includegraphics[width=0.5\linewidth]{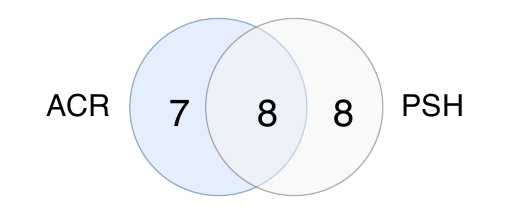}
    \caption{Overlap of detected memorization cases using PSH and ACR}
    \label{fig:psh_acr}
\end{figure}

We compare the detection results of a whitebox approach, ACR~\cite{schwarzschild_rethinking_2024} against our blackbox approach, PSH. Both are applied to Pythia 410m. Since ACR was applied by its authors on  a subset of 100 samples of FamousQuotes dataset, we execute PEARL with the text completion task on this dataset. Using the same sensitivity threshold (\(\alpha = 0.2\)) set based on the false positive rates of PEARL for RefinedWeb, we detect overall 16 cases of memorization while ACR detects 15. 
Figure~\ref{fig:psh_acr} shows that our detection overlaps on 50\% of the memorization cases identified with PSH.

\begin{figure}[!h]
    \centering
    \includegraphics[width=0.8\linewidth]{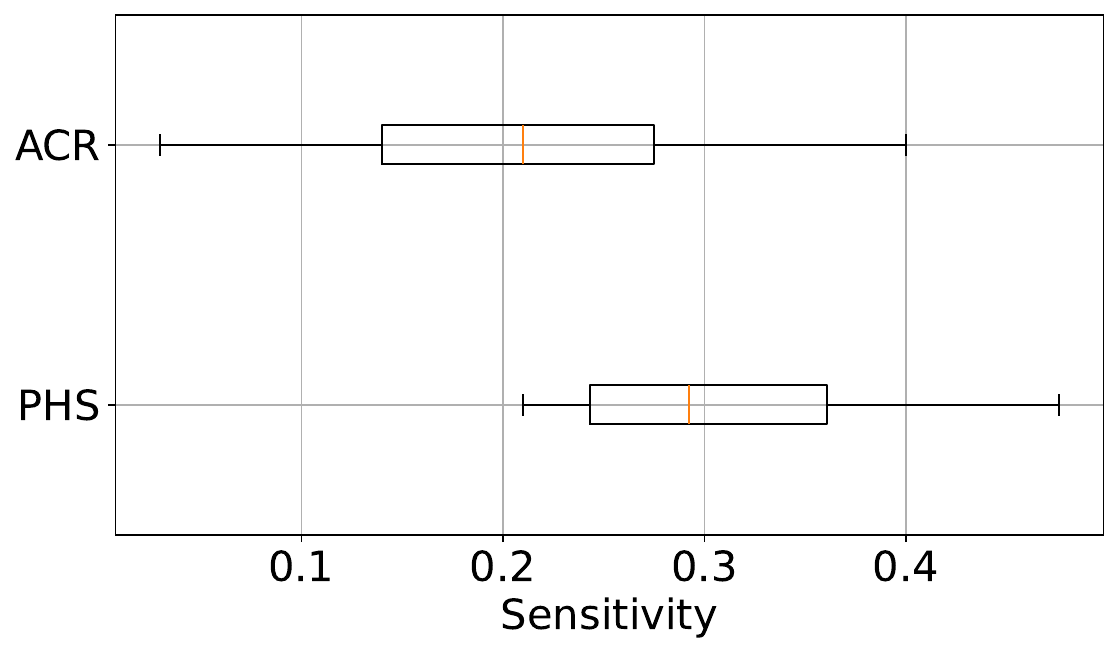}
    \caption{Distributions of sensibility values provided by PEARL for samples detected using ACR vs samples detected using PSH}
    \label{fig:psh_acr_dist}
\end{figure}

Figure~\ref{fig:psh_acr_dist} presents the distribution of sensitivity values for the samples detected through and those detected through ACR.

\subsection*{Annex 2 : Evolution of FPR given variations of the sensitivity threshold \(\alpha\) for different sizes of the Pythia model}\label{annexe:alpha_model_size}     \label{fig:model_size}

To identify the relevant sensitivity thresholds for different size-variants of Pythia, we apply them on RefineWeb and assess the FPR scores that are obtained by varying the $\alpha$ hyperparameter. The figure below provides the evolution of the FPR scores.

\begin{figure}[!htb]
    \centering
    \includegraphics[width=0.8\linewidth]{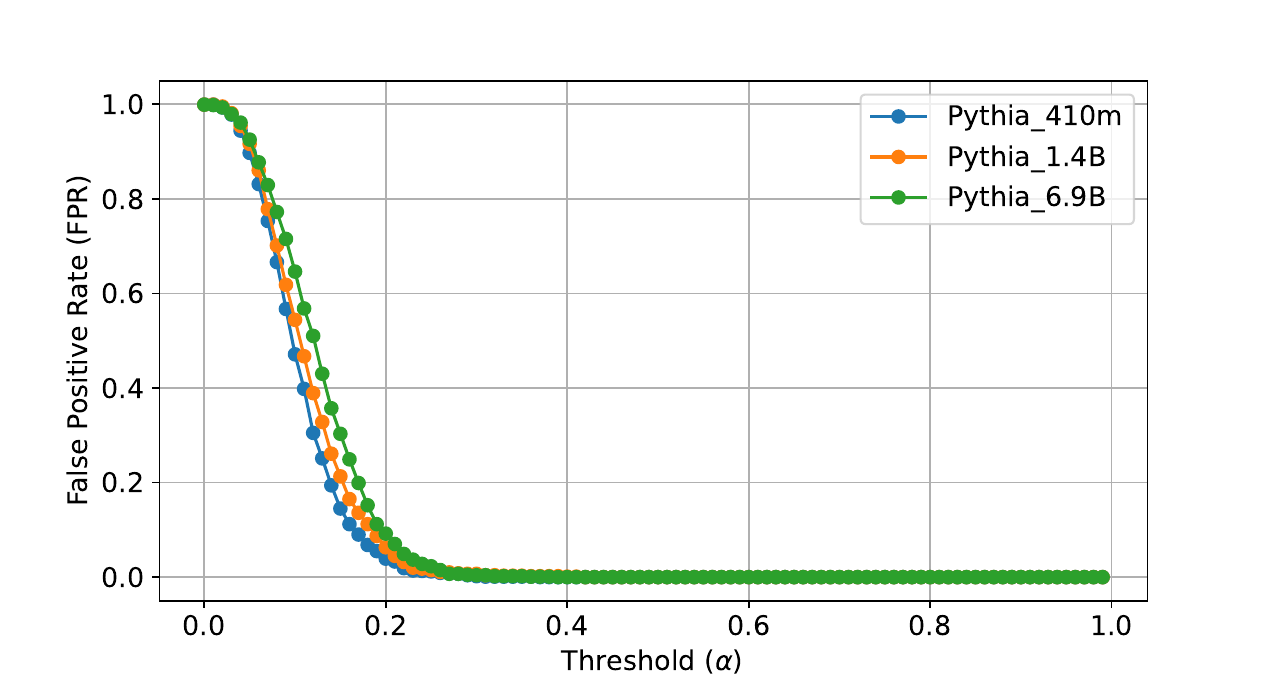}
\end{figure}

\subsection*{Annex 3 : Evolution of FPR on the LBPP dataset given variations of the sensitivity threshold $\alpha$ with the GPT\_4o model - code completion task}\label{annexe:alpha_gpt_4o_completion}
\label{fig:lim_gpt_4}

\begin{figure}[!htb]
    \centering
    \includegraphics[width=0.8\linewidth]{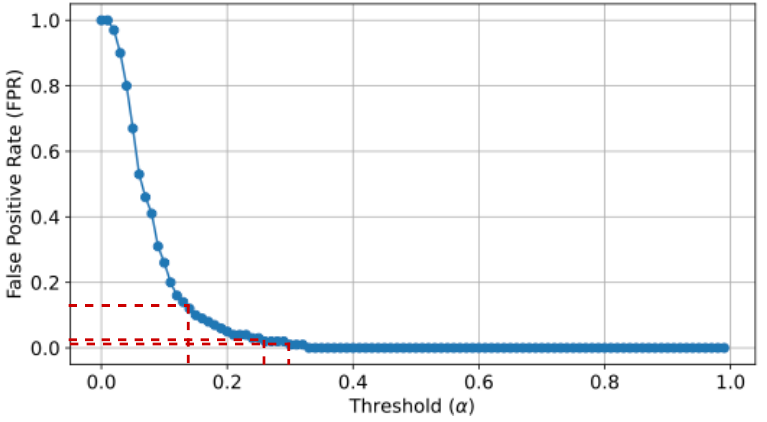}
\end{figure}

\subsection*{Annex 4 :  Evolution of FPR on the LBPP dataset given variations of the sensitivity threshold $\alpha$ with the GPT\_4o model - code summarization task}\label{annexe:alpha_gpt_4o_summary}
\label{fig:lim_summary_gpt_4}

\begin{figure}[!h]
    \centering
    \includegraphics[width=0.8\linewidth]{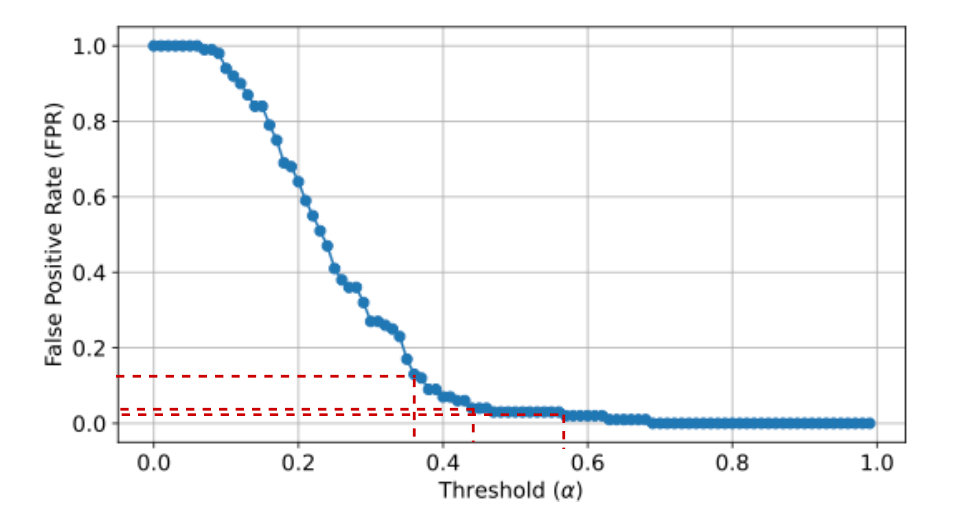}
\end{figure}

\subsection*{Annex 5 : PSH Illustration Examples} \label{annexe:1}

This annex presents the examples used to illustrate the PSH in the section \ref{sec:background}.

\highlight{Data sample in training set}

\textbf{Shakespeare's text} : Extract of Cymbeline, King of Britain (1609).\\
\textbf{Input :}
My fault being nothing—as I have told you oft— But that two villains, whose false oaths prevail'd Before my perfect honour, swore to Cymbeline I was confederate with the Romans: so Follow'd my banishment, and this twenty years This rock and these demesnes have been my world; Where I have lived at honest freedom, paid More pious debts to heaven than in all The fore-end of my time. But up to the mountains! This is not hunters' language: he that strikes The venison first shall be the lord o' the feast; To him the other two shall minister; And we will fear no poison, which attends In place of greater state. I'll meet you in the valleys. [Exeunt GUIDERIUS and ARVIRAGUS] How hard it is to hide the sparks of nature! These boys know little they are sons to the king; Nor Cymbeline dreams that they are alive. They think they are mine; and though train'd up thus meanly I' the cave wherein they bow, their thoughts do hit The roofs of palaces, and nature prompts them In simple and low things to prince it much Beyond the trick of others. This Polydore, The heir of Cymbeline and Britain, who The king his father call'd Guiderius,—Jove! When on my three-foot stool I sit and tell The warlike feats I have done, his spirits fly out Into my story: say 'Thus, mine enemy fell, And thus I set my foot on 's neck;' even then The princely blood flows in his cheek, he sweats, Strains his young nerves and puts himself in posture That acts my words. The younger brother, Cadwal, Once Arviragus, in as like a figure, Strikes life into my speech and shows much more His own conceiving.—Hark, the [...]

\textbf{Reference output:}
[...]game is roused! O Cymbeline! heaven and my conscience knows Thou didst unjustly banish me: whereon, At three and two years old, I stole these babes; Thinking to bar thee of succession, as Thou reft'st me of my lands. Euriphile, Thou wast their nurse; they took thee for their mother, And every day do honour to her grave: Myself, Belarius, that am Morgan call'd, They take for natural father. The game is up.

\highlight{Data sample outside of training set}

\textbf{BBC's article} : Extract from BBC (Published on 3rd July 2024)\\

\textbf{Input :}
A humpback whale was spotted by Jersey schoolchildren returning from a day trip to Sark. Teachers and pupils from St Lawrence Primary School and d'Auvergne School captured the sighting on video and in photos as they travelled on Monday. Donna Gicquel de Gruchy, from British Divers Marine Life Rescue Channel Islands, said the humpback whale appeared to be a young one. She said it was a "very lucky and rare sighting" and she hoped it was a sign of healthy waters. Two humpback whales were also spotted off the Channel Islands in July last year. Local wildlife expert Liz Sweet described the recent sightings of the creatures in Guernsey waters as both "very rare" and "very special". Ms Sweet said the increase could be down to the whales getting confused. She said: "There is a massive migration route through the Bay of Biscay and up around the Irish coast as they head to their feeding grounds in the Arctic. "It is really busy, really noisy water and it is easy for these animals to get a little bit confused, maybe a little bit lost." Her [...] \\

\textbf{Reference output :}
[...]other theory was that rising sea temperatures could be luring the whales' food closer to the Channel Islands. Ms Sweet explained: "Fish are moving around a lot more so it might have been following food so it could have just been here for a stop off.

\subsection*{Annex 6 : List of identified GPT\_4o's memorisation instances} \label{annexe:2}
This annex present the data identified as memorised regarding the completion task describe in the section \ref{sec:real_case} with \(\alpha = 0.29\) summarize in the table \ref{tab:list_memorized}.
The column \textit{Problem ref} identify the problem in it original dataset.

The instances of Bible's text that we identified as memorized are the following : 

\textit{\#262:}O thou that hearest prayer, unto thee shall all flesh come.Iniquities prevail against me: as for our transgressions, thou shalt purge them away.Blessed is the man whom thou choosest, and causest to approach unto thee, that he may dwell in thy courts: we shall be satisfied with the goodness of thy house, even of thy holy temple.By terrible things in righteousness wilt thou answer us, O God of our salvation; who art the confidence of all the ends of the earth, and of them that are afar off upon the sea:Which by his strength setteth fast the mountains; being girded with power:Which stilleth the noise of the seas, the noise of their waves, and the tumult of the people.They also that dwell in the uttermost parts are afraid at thy tokens: thou makest the outgoings of the morning and evening to rejoice.Thou visitest the earth, and waterest it: thou greatly enrichest it with the river of God, which is full of water: thou preparest them corn, when thou hast so provided for it.Thou waterest the ridges thereof abundantly: thou settlest the furrows thereof: thou makest it soft with showers: thou blessest the springing thereof.Thou crownest the year with thy goodness; and thy paths drop fatness.They drop upon the pastures of the wilderness: and the little hills rejoice on every side.The pastures are clothed with flocks; the valleys also are covered over with corn; they shout for joy, they also sing.Make a joyful noise unto God, all ye lands:Sing forth the honour of his name: make his praise glorious. \\

\textit{\#289:}But let every man prove his own work, and then shall he have rejoicing in himself alone, and not in another.For every man shall bear his own burden.Let him that is taught in the word communicate unto him that teacheth in all good things.Be not deceived; God is not mocked: for whatsoever a man soweth, that shall he also reap.For he that soweth to his flesh shall of the flesh reap corruption; but he that soweth to the Spirit shall of the Spirit reap life everlasting.And let us not be weary in well doing: for in due season we shall reap, if we faint not.As we have therefore opportunity, let us do good unto all men, especially unto them who are of the household of faith.Ye see how large a letter I have written unto you with mine own hand.As many as desire to make a fair shew in the flesh, they constrain you to be circumcised; only lest they should suffer persecution for the cross of Christ.For neither they themselves who are circumcised keep the law; but desire to have you circumcised, that they may glory in your flesh.But God forbid that I should glory, save in the cross of our Lord Jesus Christ, by whom the world is crucified unto me, and I unto the world.For in Christ Jesus neither circumcision availeth any thing, nor uncircumcision, but a new creature.And as many as walk according to this rule, peace be on them, and mercy, and upon the Israel of God. \\

\begin{table*}[t]
    \centering
    \scalebox{0.9}{
    \begin{tabular}{c|c|c|c|c|c|c|c|c|c}
         \#ID & Source & \(m(Y^*_0)\) & \(m(Y^*_1)\) & \(m(Y^*_2)\) & \(m(Y^*_3)\) & \(m(Y^*_4)\) & \(m(Y^*_5)\) & sensitivity & Problem ref \\
         \hline
         \#17 & LBPP & 0.42 & 0.23 & 0.51 & 0.18 & 0.21 & 0.25 & 0.32 & LBPP/17 \\ 
         \#58 & LBPP & 0.32 & 0.62 &	0.37 &	0.35 &	0.34 &	0.45 &	0.29 & LBPP/58 \\ 
         \#103 & HumanEval & 0.8 & 0.49 & 0.8 & 0.73 & 0.38 & 0.24& 0.34 & HumanEval/3 \\
        \#105 & HumanEval & 0.68 & 0.32 & 0.42 & 0.55 & 0.32 & 0.24& 0.36 & HumanEval/5 \\
        \#107 & HumanEval & 0.66 & 0.24 & 0.21 & 0.19 & 0.14 & 0.12& 0.42 & HumanEval/7 \\
        \#113 & HumanEval & 0.71 & 0.53 & 0.24 & 0.76 & 0.3 & 0.26& 0.53 & HumanEval/13 \\
        \#114 & HumanEval & 0.55 & 0.56 & 0.27 & 0.49 & 0.29 & 0.22& 0.3 & HumanEval/14 \\
        \#115 & HumanEval & 0.65 & 0.3 & 0.26 & 0.29 & 0.17 & 0.24& 0.35 & HumanEval/15 \\
        \#116 & HumanEval & 0.62 & 0.38 & 0.2 & 0.16 & 0.44 & 0.12& 0.32 & HumanEval/16 \\
        \#122 & HumanEval & 0.67 & 0.33 & 0.21 & 0.67 & 0.19 & 0.19& 0.48 & HumanEval/22 \\
        \#123 & HumanEval & 0.87 & 0.68 & 0.22 & 0.19 & 0.21 & 0.08& 0.46 & HumanEval/23 \\
        \#127 & HumanEval & 0.56 & 0.19 & 0.72 & 0.19 & 0.13 & 0.14& 0.53 & HumanEval/27 \\
        \#128 & HumanEval & 0.64 & 0.2 & 0.41 & 0.22 & 0.26 & 0.17& 0.44 & HumanEval/28 \\
        \#129 & HumanEval & 0.71 & 0.33 & 0.18 & 0.2 & 0.62 & 0.18& 0.44 & HumanEval/29 \\
        \#130 & HumanEval & 0.55 & 0.15 & 0.18 & 0.13 & 0.17 & 0.15& 0.4 & HumanEval/30 \\
        \#134 & HumanEval & 0.66 & 0.24 & 0.22 & 0.08 & 0.15 & 0.15& 0.43 & HumanEval/34 \\
        \#142 & HumanEval & 0.55 & 0.49 & 0.42 & 0.13 & 0.17 & 0.16& 0.29 & HumanEval/42 \\
        \#145 & HumanEval & 0.46 & 0.16 & 0.29 & 0.11 & 0.09 & 0.11& 0.3 & HumanEval/45 \\
        \#154 & HumanEval & 0.67 & 0.14 & 0.41 & 0.14 & 0.18 & 0.1& 0.53 & HumanEval/54 \\
        \#162 & HumanEval & 0.49 & 0.49 & 0.12 & 0.16 & 0.15 & 0.16& 0.37 & HumanEval/62 \\
        \#166 & HumanEval & 0.64 & 0.57 & 0.2 & 0.35 & 0.22 & 0.39& 0.37 & HumanEval/66 \\
        \#175 & HumanEval & 0.29 & 0.43 & 0.13 & 0.23 & 0.14 & 0.23& 0.3 & HumanEval/75 \\
        \#180 & HumanEval & 0.7 & 0.67 & 0.71 & 0.29 & 0.34 & 0.33& 0.42 & HumanEval/80 \\
        \#188 & HumanEval & 0.43 & 0.13 & 0.24 & 0.16 & 0.21 & 0.15& 0.29 & HumanEval/88 \\
        \#198 & HumanEval & 0.67 & 0.61 & 0.32 & 0.31 & 0.34 & 0.35& 0.29 & HumanEval/98 \\
         \#262 & Bible & 0.65 & 0.18 & 0.18 & 0.09 & 0.1 & 0.09& 0.47 & - \\
        \#289 & Bible & 0.63 & 0.23 & 0.09 & 0.09 & 0.08 & 0.08& 0.4 & - \\

    \end{tabular}
    }
    \caption{List of identified memorized cases in GPT\_4o regarding the text completion task.}
    \label{tab:list_memorized}
\end{table*}

\subsection*{Annex 7 : Comparaison sensitivity between NY Times and Bible's text} \label{annexe:low_high_sensitivity}

This annex presents the instances that present the highest sensitivity and the lowest one regarding the NY Times and Bible's text by considering the completion task introduced in the section \ref{sec:real_case}. The result is summarized in the table \ref{tab:extrem_sensitivity_value}

\begin{table}[h]
    \centering
    \begin{tabular}{c|c|c|c|c}
        Source & \multicolumn{2}{|c}{Lowest sensitivity} & \multicolumn{2}{|c}{Highest sensitivity} \\
        \hline
          & \#ID & Sensitivity & \#ID & Sensitivity \\
          \hline
          Bible & \#261 & 0.019 & \#262 & 0.469 \\
          \hline
          NY Times & \#301 & 0.007 & \#304 & 0.266 \\
          \hline
          
    \end{tabular}
    \caption{Lowest and highest sensitivity in }
    \label{tab:extrem_sensitivity_value}
\end{table}

\textit{\#261:}Now the sons of Reuben the firstborn of Israel, (for he was the firstborn; but forasmuch as he defiled his father's bed, his birthright was given unto the sons of Joseph the son of Israel: and the genealogy is not to be reckoned after the birthright.For Judah prevailed above his brethren, and of him came the chief ruler; but the birthright was Joseph's:)The sons, I say, of Reuben the firstborn of Israel were, Hanoch, and Pallu, Hezron, and Carmi.The sons of Joel; Shemaiah his son, Gog his son, Shimei his son,Micah his son, Reaia his son, Baal his son,Beerah his son, whom Tilgathpilneser king of Assyria carried away captive: he was prince of the Reubenites.And his brethren by their families, when the genealogy of their generations was reckoned, were the chief, Jeiel, and Zechariah,And Bela the son of Azaz, the son of Shema, the son of Joel, who dwelt in Aroer, even unto Nebo and Baalmeon:And eastward he inhabited unto the entering in of the wilderness from the river Euphrates: because their cattle were multiplied in the land of Gilead.And in the days of Saul they made war with the Hagarites, who fell by their hand: and they dwelt in their tents throughout all the east land of Gilead.And the children of Gad dwelt over against them, in the land of Bashan unto Salcah:Joel the chief, and Shapham the next, and Jaanai, and Shaphat in Bashan.And their brethren of the house of their fathers were, Michael, and Meshullam, and Sheba, and Jorai, and Jachan, and Zia, and Heber, seven.

\textit{\#262:}O thou that hearest prayer, unto thee shall all flesh come.Iniquities prevail against me: as for our transgressions, thou shalt purge them away.Blessed is the man whom thou choosest, and causest to approach unto thee, that he may dwell in thy courts: we shall be satisfied with the goodness of thy house, even of thy holy temple.By terrible things in righteousness wilt thou answer us, O God of our salvation; who art the confidence of all the ends of the earth, and of them that are afar off upon the sea:Which by his strength setteth fast the mountains; being girded with power:Which stilleth the noise of the seas, the noise of their waves, and the tumult of the people.They also that dwell in the uttermost parts are afraid at thy tokens: thou makest the outgoings of the morning and evening to rejoice.Thou visitest the earth, and waterest it: thou greatly enrichest it with the river of God, which is full of water: thou preparest them corn, when thou hast so provided for it.Thou waterest the ridges thereof abundantly: thou settlest the furrows thereof: thou makest it soft with showers: thou blessest the springing thereof.Thou crownest the year with thy goodness; and thy paths drop fatness.They drop upon the pastures of the wilderness: and the little hills rejoice on every side.The pastures are clothed with flocks; the valleys also are covered over with corn; they shout for joy, they also sing.Make a joyful noise unto God, all ye lands:Sing forth the honour of his name: make his praise glorious.

\textit{\#301:}I think my credibility would be affected if I didn’t ask experts for their opinion,” he said.  The governor also said that the discrepancy between the predictions and the actual statistics was because of the behavior of New Yorkers themselves. With some exceptions, New Yorkers have managed to follow the restrictions on movement and socializing.  Dr. Deborah Birx, the White House coronavirus response coordinator, seemed to agree and congratulated Mr. Cuomo and his counterparts on Friday for having slowed the tide of infections in their states.  “That has dramatically changed because of the impact of what the citizens of New York and New Jersey and across Connecticut and now Rhode Island are doing to really change the course of this pandemic,” Dr. Birx said.  It is, of course, prudent politically and for public health reasons that elected leaders over-plan, not under-plan, for disasters. During hurricanes, for instance, governors are much more likely to save their constituents’ lives, and their own jobs, by ordering evacuations early rather than banking on the chance that a storm will peter out.  The main objective in “flattening the curve” of the outbreak, apart from keeping people from dying, is to slow the spread enough to keep hospitals functioning.  And from the start of the coronavirus emergency, Mr. Cuomo has repeatedly taken the position that he would rather be prepared for a dire scenario that never came to pass than to blithely put his faith in optimistic forecasts. Ironically, his doomsday attitude may complicate his efforts to keep the state on course as New Yorkers start to realize that the worst has not happened and eventually get itchy to go out.

\textit{\#304:}--Monkey Eats From Bird Feeder After Escaping Scottish Wildlife Park   --French Farmers Block Roads Around Paris, Escalating Protests   --Russian Military Plane Crashes Near Border With Ukraine --Winter Storm Makes Landings Difficult at Heathrow   --U.S. Pressed Israel to Reduce Civilian Suffering in Gaza, Blinken Says  Recent episodes in Europe Historic Copenhagen Stock Exchange Partly Collapses in Fire  --Historic Copenhagen Stock Exchange Partly Collapses in Fire One Person Is Killed in a Cable Car Accident in Turkey   --One Person Is Killed in a Cable Car Accident in Turkey German Police Stop Pro-Palestinian Conference  --German Police Stop Pro-Palestinian Conference Smoke Rings Rise From Mt. Etna  --Smoke Rings Rise From Mt. Etna Nightclub Fire in Istanbul Kills 29 People  0:48 Nightclub Fire in Istanbul Kills 29 People Waiters Compete in Paris’ Revived Cafe Race  --Waiters Compete in Paris’ Revived Cafe Race Princess of Wales Announces Cancer Diagnosis  --Princess of Wales Announces Cancer Diagnosis Russian Strikes Cut Off Electricity and Disrupt Water Supply in Kharkiv   --Russian Strikes Cut Off Electricity and Disrupt Water Supply in Kharkiv Homes Are Destroyed by Russian Attack in Southeastern Ukraine   --Homes Are Destroyed by Russian Attack in Southeastern Ukraine Missile Attack on Kyiv  --Missile Attack on Kyiv Volcano Erupts in Southwestern Iceland  --Volcano Erupts in Southwestern Iceland France Enshrines Abortion Rights in Its Constitution  --France Enshrines Abortion Rights in Its Constitution.